\documentclass[12pt]{article}

\usepackage{newtxtext,newtxmath}
\usepackage{graphicx}

\usepackage[letterpaper,margin=1in]{geometry}

\renewenvironment{abstract}
	{\quotation}
	{\endquotation}

\date{}

\makeatletter
\renewcommand{\fnum@figure}{\textbf{Figure \thefigure}}
\renewcommand{\fnum@table}{\textbf{Table \thetable}}
\makeatother

\usepackage{scicite}

\usepackage{url}

\def\scititle{
A compact neuromorphic system for ultra-energy-efficient, on-device robot localization
}
\title{\bfseries \boldmath \scititle}

\author
{Adam D Hines$^{1\ast}$, Michael Milford$^{1}$, Tobias Fischer$^{1}$\and
\small{$^{1}$QUT Centre for Robotics, Queensland University of Technology, 2 George St, Brisbane, QLD 4001, Australia}\and
\small{$^\ast$Corresponding author. Email:  adam.hines@qut.edu.au}
}

\usepackage{eso-pic}
\usepackage{url}
\AddToShipoutPicture*{%
     \AtTextUpperLeft{%
         \put(-3.5,10){
           \begin{minipage}{\textwidth}
              \scriptsize
              \MakeUppercase{This is the author's version of the work. It is posted here by permission of the AAAS for personal use, not for redistribution. The definitive version was published in Science Robotics on 06/18/2025, DOI:} \url{https://doi.org/10.1126/scirobotics.ads3968.}
           \end{minipage}}%
     }%
}

\usepackage{float}

\begin{document} 

\maketitle 

\begin{abstract} \bfseries \boldmath
Neuromorphic computing offers a transformative pathway to overcome the computational and energy challenges faced in deploying robotic localization and navigation systems at the edge. Visual place recognition, a critical component for navigation, is often hampered by the high resource demands of conventional systems, making them unsuitable for small-scale robotic platforms which still require accurate long-endurance localization. Although neuromorphic approaches offer potential for greater efficiency, real-time edge deployment remains constrained by the complexity of bio-realistic networks. In order to overcome this challenge, fusion of hardware and algorithms is critical to employ this specialized computing paradigm. Here, we demonstrate a neuromorphic localization system that performs competitive place recognition in up to 8 kilometers of traversal using models as small as 180 kilobytes with 44,000 parameters, while consuming less than 8\% of the energy required by conventional methods. Our Locational Encoding with Neuromorphic Systems (LENS) integrates spiking neural networks, an event-based dynamic vision sensor, and a neuromorphic processor within a single SynSense Speck\texttrademark{} chip, enabling real-time, energy-efficient localization on a hexapod robot. When compared to a benchmark place recognition method, Sum-of-Absolute-Differences (SAD), LENS performs comparably in overall precision. LENS represents an accurate fully neuromorphic localization system capable of large-scale, on-device deployment for energy efficient robotic place recognition.
Neuromorphic computing enables resource-constrained robots to perform energy efficient, accurate localization.
\end{abstract}

\section*{Introduction}

Robot localization is a critical component of many autonomous navigation systems, enabling robots to determine their location whilst supporting the ability to understand and interact with their surroundings. A central challenge in robot localization is visual place recognition (VPR), which requires robots to identify and classify previously visited locations under varying conditions~\cite{Schubert2023,Zhang2021,Masone2021}. Conventional VPR strategies often rely on deep convolutional neural networks or transformer-based architectures to robustly extract features from the environment~\cite{Berton2022,Neuhold2017,Cordts2016,Izquierdo2024}. For resource-constrained robotic platforms that nonetheless require the ability to keep track of where they are located over vast distances, using such conventional VPR methods is often not feasible due to their computational demands. An open challenge for real-world deployment of VPR is to find models that are computationally and energy efficient, and can be deployed at the edge.

Neuromorphic computing, which takes inspiration by the brain, has emerged as a promising solution for addressing the energy and computational challenges associated with robot localization and navigation~\cite{Orchard2021,Painkras2012,Zhu2023-2,vanDijk2024,Yu2023}. The human brain's ability to learn and navigate complex environments efficiently~\cite{Friston2010,Chen2024,Lisman2013,Liu2023} has inspired roboticists to create more computationally efficient localization systems~\cite{Milford2004,Yu2019,Wang2023,Wang2012,Joseph2023}. In particular, Spiking Neural Networks (SNNs) have been widely explored due to their ability to perform tasks through bio-realistic simulations of neuron activity~\cite{Tavanaei2019,Eshraghian2023}. Although several SNN models have been proposed for robotic localization~\cite{Hussaini2023,Hussaini2022,Hussaini2023-2,Hines2024,Yu2023,Zhu2023-2,Giraldo2023,Liu2021,Tang2019,Galluppi2012,Bi1998}, they ultimately have not yet fully lived up to the promise of computationally efficient deployment due to the complexity of continually modeling neurons in real-time~\cite{Hussaini2023-2,Hussaini2023,Wang2023}.

To advance the practicality of neuromorphic computing in VPR, the focus must shift toward developing models that can be effectively deployed on Size, Weight, and Power (SWaP) constrained robotic platforms. One approach to overcoming deployment barriers is to trade the intricate bio-realism of traditional SNNs for simpler, more efficient networks that still deliver robust performance~\cite{Chancan2020}. However, small-scale systems can often be limited in their ability to map large environments, which restricts the downstream use-cases~\cite{vanDijk2024}. To improve efficiency, SNNs full potential for robotic deployment can be realized by integrating them with neuromorphic hardware~\cite{Bartolozzi2022,Gallego2022,vanDijk2024,Yu2023,Wang2023,Russo2023,Yang2023,Zhu2023-2} such as Intel\textsuperscript{\textregistered}'s Loihi 2, TrueNorth, and Tianjic~\cite{Orchard2021,Akopyan2015, Pei2019}, which draw inspiration from neuroscience for their design, enabling them to transmit and receive physical spikes from neurons within the processor cores~\cite{Shastri2018}. Neuromorphic computing has been successfully utilized in tasks other than localization, such as control systems for drone flight~\cite{Paredes-Valles2024}, route following~\cite{vanDijk2024}, autonomous driving controllers~\cite{Halay2023}, and motor controllers~\cite{Linares-Barranco2020}. In contrast, non-spiking algorithms that are brain-inspired are widely used in localization tasks  but have limited ability to be deployed on neuromorphic hardware. Although several neuromorphic localization systems have been developed and tested, they are limited by environment scale~\cite{vanDijk2024}, require complex multi-modal fusion of large network models~\cite{Yu2023}, or have restricted real-time capability~\cite{Zhu2023-2}.

Event cameras, including Dynamic Vision Sensors (DVS), offer further computational advantages when paired with neuromorphic processors as they only transmit information based on pixel-wise changes in light intensity exceeding a threshold, thereby reducing unnecessary data processing~\cite{Gallego2022}. In the context of SNNs, spiking activity can easily be triggered and propagated based on incoming input event streams and are well suited to process them. Fusing neuromorphic algorithms, sensors, and processors to perform robotic localization tasks therefore provides a promising avenue to overcome the computational limitations inherent in SNN-based VPR~\cite{Zhu2023-2,Yu2023}.

\begin{figure}[H]
    \includegraphics[width=\textwidth]{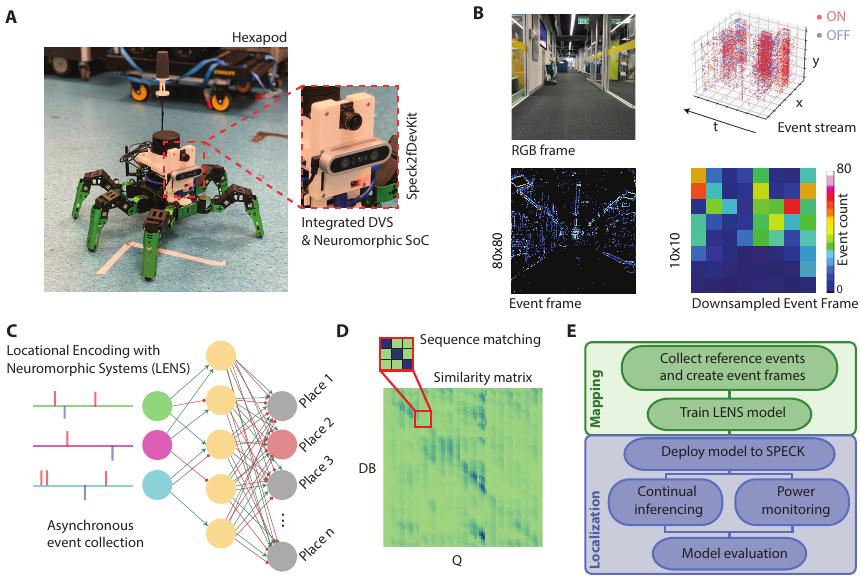}
    \caption{\textbf{Neuromorphic place recognition system on a hexapod robot.} (\textbf{A}) Our system was deployed on a HiWonder Jetahexa hexapod robotic platform equipped with a custom 3D-printed mount housing a SynSense Speck2fDevKit. Event streams captured by the onboard DVS are processed into event frames by counting ON and OFF events during the hexapod's traversal (\textbf{B}). The initial $128\times128$ input is cropped to a $80\times80$ region of interest which is then downsampled to $10\times10$ by selecting central pixels via a 2D convolutional layer (see Materials and Methods: Event processing). (\textbf{C}) The spiking neural network, Locational Encoding with Neuromorphic Systems (LENS), learns reference event frames for on-chip deployment, enabling real-time localization using asynchronously collected events. (\textbf{D}) Sequence matching techniques enhance the similarity between reference and query inputs, improving overall precision. (\textbf{E}) Schematic of the fully neuromorphic visual place recognition pipeline.}
    \label{fig:schemafig}
    \end{figure}

Here, we introduce a neuromorphic pipeline for robotic localization that is compact, real-time capable, and able to map environments up to 8~km in length. Our system, Locational Encoding with Neuromorphic Systems (LENS), is designed to deliver high accuracy and efficiency with a model size of less than 180~KB and 44~K parameters, capable of processing up to 8~km of traversal data while consuming less than 8\% of the energy required by conventional platforms. We train a SNN using a temporal time to first spike encoding scheme designed to perform VPR~\cite{Hines2024} and deploy our model on the SynSense Speck\texttrademark{}~\cite{Yao2024}, which combines a DVS and system-on-chip neuromorphic processor. The LENS neuromorphic algorithm was developed in tandem with neuromorphic hardware deployment to deliver ultra energy-efficient place recognition for robotic localization.  By deploying on Speck\texttrademark{}, we forgo the need for any external sensory modalities or computational resources as all localization is performed on-chip. By abstracting complex bio-realism in favor of enhanced performance, LENS represents a notable advancement in neuromorphic localization, as a fully event-driven platform designed for VPR. We validate LENS on a HiWonder JetHexa hexapod robot (Fig.~\ref{fig:schemafig}), demonstrating its effectiveness in both indoor and outdoor small-scale traversals ($\approx$25-40~m, 50-80 places) and large-scale datasets ($\approx$8~km, $\>$600 places)~\cite{Fischer2020}, showcasing its potential for online VPR.

\section*{Results}

\begin{figure}[H]
    \centering
    \includegraphics[width=0.83\textwidth]{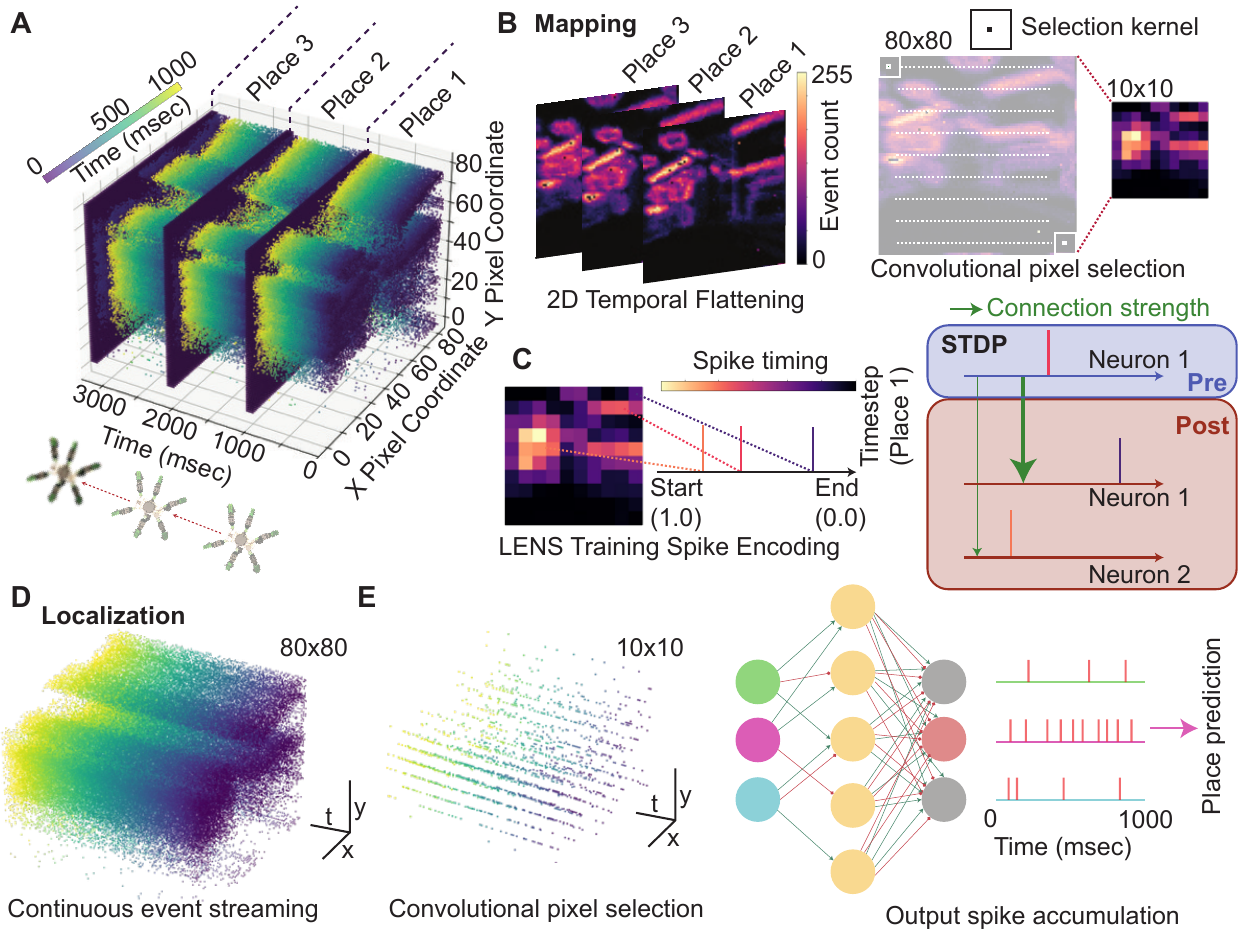}
    \caption{\textbf{Event stream processing using LENS.} (\textbf{A}) Event streams are collected from the Spec2fDevKit board from an $80\times80$ region of interest using a DAVIS128 sensor as our hexapod robot moves through the environment. One place represents one second or $\approx$0.15$\,\text{m}$ of movement. The event stream is then used in two unique ways during mapping and localization phases. (\textbf{B}) During mapping, events over the one second time period are counted and flattened into a 2D-temporal representation of the place. We use a convolutional selection kernel to pick the center pixel from the 2D image (see Materials and Methods: Model training), resulting in a $10\times10$ down sampled image. (\textbf{C}) Down sampled images are normalized in the range of $[0, 1]$ with pixel intensity determining spike timing in our LENS training regime, following~\cite{Hines2024}. Using spike-timing dependent plasticity (STDP) across the network layers, we strengthen or weaken both excitatory and inhibitory connections if the spike occurs after or before pre-synaptic activity, respectively. (\textbf{D}) During localization, we forgo the temporal representation of places and use the raw event streams to continuously activate and bias the LENS network on-chip. (\textbf{E}) We apply the same convolutional pixel selection as the first layer of our neural network, deployed on the Speck\texttrademark{} directly. Over the time collection period of one second, we accumulate output spikes from our model and make a place prediction based on the neuron with the highest spiking activity.}
    \label{fig:lensschema}
    \end{figure}

\paragraph*{Locational encoding using neuromorphic systems}
An overview of our Locational Encoding using Neuromorphic Systems (LENS) network is presented in Figure~\ref{fig:lensschema}. We used event streams from the Speck\texttrademark{} DAVIS128 sensor in two different ways for the mapping and localization phases of the robot navigation. A place was represented as one second of movement, translating to a distance of $\approx$0.15$\,\text{m}$ (Fig.~\ref{fig:lensschema}A). For the mapping phase, the number of events were counted per pixel and then flattened into a 2D-temporal representation image of the place (Fig.~\ref{fig:lensschema}B). Using a 2D-convolutional kernel with the center weight set to 1 and all other weights set to 0 (see Materials and Methods: Model training), we selected the center pixel from the 2D image to downsample to $10\times10$ pixels to fit in our LENS system (Fig.~\ref{fig:lensschema}B). Following Hines et al.~\cite{Hines2024}, we trained our model by first normalizing pixel intensities between [0, 1] to use a temporally coded spike scheme such that higher pixel intensities result in earlier spiking with lower intensities activating later (Fig.~\ref{fig:lensschema}C, see Materials and Methods: Model training). Connection weights were trained via spike-timing dependent plasticity learning, with enhanced connectivity when post-synaptic activity occurred after pre-synaptic spikes and diminished connectivity if post-synaptic neurons spike beforehand (Fig.~\ref{fig:lensschema}C, see Materials and Methods: Model training). For our three layer architecture, we used unsupervised learning between the input and feature layers and a supervised delta learning rule between the feature and output layer (see Materials and Methods: Model training). 

During the localization phase, we used the raw event streams collected from the Speck\texttrademark{} directly into our model instead of 2D-temporal representations of places (Fig.~\ref{fig:lensschema}D, see Materials and Methods: Deployment time). The first layer of our LENS system was the convolutional pixel selection filter applied earlier to reduce the input dimensionality to $10\times10$ pixels (Fig.~\ref{fig:lensschema}E, see Materials and Methods: Deployment time). We performed place recognition with asynchronous events over one second collection windows and picked the location with the highest spiking activity for the neuron in the output layer (Fig.~\ref{fig:lensschema}E, see Materials and Methods: Online place matching).  

\paragraph*{Power and energy efficiency}

\begin{figure}[H]
    \includegraphics[width=\textwidth]{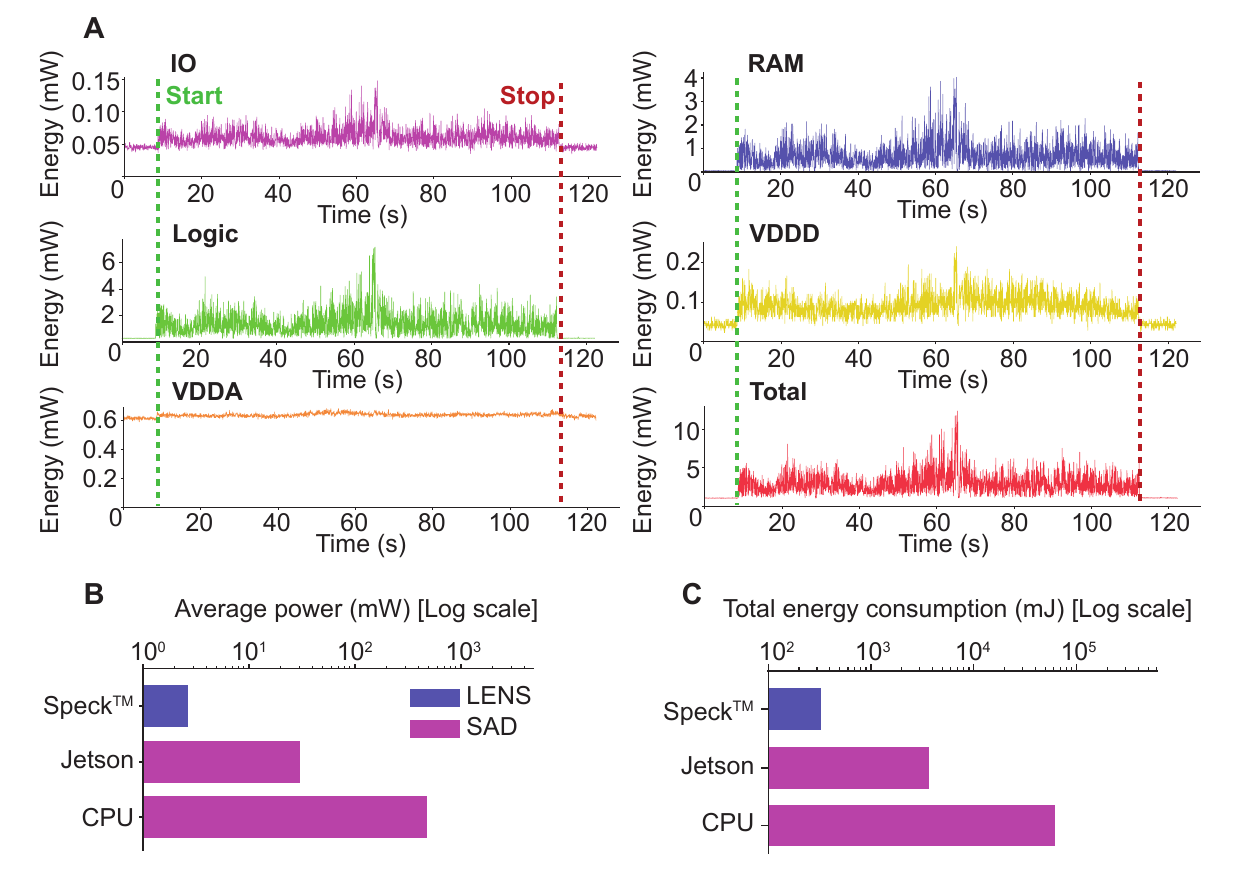}
    \caption{\textbf{Energy efficiency of Real-time VPR on a neuromorphic processor vs.~von Neumann hardware.} (\textbf{A}) Power consumption was recorded across 5 components of the Speck\texttrademark{} neuromorphic processor: input/output (IO), memory (RAM), logic, digital components (VDDD), analog components (VDDA), and the total. Start indicates when the robot began to move until the navigation task has finished, indicated by the Stop. (\textbf{B}). The Speck\texttrademark{} processor consumed substantially less total energy compared to the Jetson Nano and Intel\textsuperscript{\textregistered} CPU (\textbf{C}). The mW and mJ values are presented on a logarithmic scale for display purposes.}
    \label{fig:energy}
    \end{figure}

Neuromorphic computing platforms offer an advantage in terms of power and energy consumption when compared to traditional von Neumann hardware, particularly in tasks like Visual Place Recognition (VPR) where energy efficiency is crucial for long-term operation on battery-powered robots. We measured the power consumed by various components of the Speck\texttrademark{} neuromorphic processor while performing VPR (Fig.~\ref{fig:energy}A). Notably, when the robot was stationary, the DVS generated very few events which resulted in further reductions in power consumption. To quantify the energy savings of LENS deployed on the Speck\texttrademark{} processor relative to von Neumann hardware, we performed VPR using the Sum-of-absolute-differences (SAD) place matching method~\cite{Milford2012} on an Intel\textsuperscript{\textregistered} i7-9700K CPU and a NVIDIA Jetson Nano, which ran the hexapod (see Materials and Methods: Power measurements). After subtracting baseline CPU power use, we found that the Speck\texttrademark{} consumed an average of 2.7~mW, which is just 8.8\% and 0.5\% of the power required by the Jetson Nano and Intel\textsuperscript{\textregistered} CPU to run SAD, respectively (Fig.~\ref{fig:energy}B).

The overall energy consumption (the integral of power over time) of the Speck\texttrademark{} was 327~mJ, with the Jetson using 2,968~mJ and the Intel\textsuperscript{\textregistered} CPU requiring 61,427~mJ, meaning overall our system required just 8.9\% and 0.5\% of the energy needed to perform the same task on von Neumann hardware (Fig.~\ref{fig:energy}C). Fewer pixel representations of places still resulted in energy consumption far greater than what was required for LENS on Speck\texttrademark{} (see Table~\ref{tab:sadenergy}). We also observed a prominent reduction in power and energy consumption deploying to Speck\texttrademark{} when running our LENS system on von Neumann hardware (Fig.~\ref{fig:lensvonneu}).

\paragraph*{Large-scale place recognition with compact models}

\begin{figure}[H]
    \centering
    \includegraphics[width=0.8\textwidth]{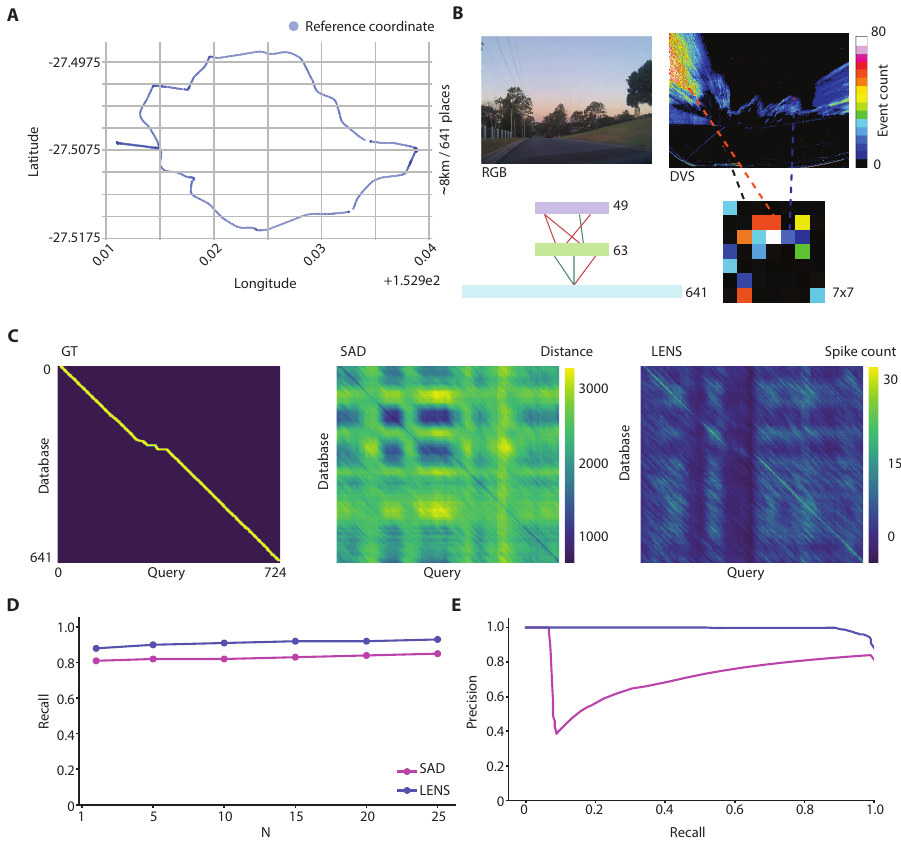}
    \caption{\textbf{Large scale place recognition using a compact neuromorphic ecosystem.} (\textbf{A}) Coordinate map of the Brisbane Event VPR dataset route~\cite{Fischer2020} which represents an $\approx$8~km traversal and 641 unique place locations around the suburbs of Brisbane, Australia. (\textbf{B}) Event frames are generated in the same way as on robot experiments (Fig.~\ref{fig:matching}). Pixels were selected randomly rather than by applying a 2D convolutional filter due to a larger event camera resolution used in~\cite{Fischer2020}. We selected pixels to generate event frames that are $7\times7$ in size. Our model architecture was modified to $49\times63\times641$ to allow for on-chip localization. (\textbf{C}) Similarity matrices for both SAD and LENS, showing how the similarity for both methods aligns closely with the ground truth (GT). (\textbf{D}) Recall@N and (\textbf{E}) Precision-Recall curves for our method when compared to SAD. LENS performed best overall compared to SAD, and also achieved better precision at lower recall indicating our system is more confident at the places it selected. The energy efficiency and compact model size of just 753 neurons with 44K parameters highlights the benefit of our system.}
    \label{fig:largevpr}
    \end{figure}

To assess our system's suitability for real-world scenarios, we evaluated its performance in large scale environments using the Brisbane Event VPR dataset containing event streams from an $\approx$8~km route in Brisbane, Australia~\cite{Fischer2020} (Fig.~\ref{fig:largevpr}). This dataset was representative of a stable, four wheeled robotic platform that our model could be deployed on. We generated 641 event frames of unique places by sampling over one second intervals from the dataset (Fig.~\ref{fig:largevpr}B). To fit the model onto the Speck\texttrademark{} chip, we used a network architecture of 49 input neurons, 63 feature layer neurons, and 641 output neurons (Fig.~\ref{fig:largevpr}B). This was the largest and highest performing model architecture we could use to run our model on-chip, balancing the number of input pixels and feature representation (Fig.~\ref{fig:modelsize}). We used the ``sunset2'' traverse for mapping (reference dataset), and the ``sunset1'' traverse for localization (query dataset). We also employed the Sum-of-absolute-difference (SAD method)~\cite{Fischer2022,Milford2012} as a baseline comparison method, calculating pixel-wise similarity between mapping (training) and localization (test) data.  

We used spikes from the model's output layer and the Euclidean distance calculated in SAD to generate similarity matrices and used sequence matching which preserves the distances of sequences matches while nullifying cross-distances, improving localization precision during deployment time (Fig.~\ref{fig:largevpr}C, see Materials and Methods: Online place matching). The similarity matrix was used to compute Recall@N (the proportion of true place matches retrieved within the top N predicted matches) and precision and recall metrics, which evaluated how well the robot is correctly identifying its location as it moves through the environment. Although other methods for event based place recognition exist, they require the reconstruction of images from event streams for deep feature extraction~\cite{Lee2021, Lee2023}. As such, we elected to not compare our system to these methods since they require higher complexity and computational time than what is feasible to run on our robotic platform.

Both LENS and SAD generated similarity matrices that closely aligned with the ground truth (Fig.~\ref{fig:largevpr}C). LENS performed best overall for Recall@N, achieving a Recall@1 of 0.88 compared to SAD's 0.81 (Fig.~\ref{fig:largevpr}D). LENS also demonstrated better precision at lower recall, indicating higher confidence in its place recognition (Fig.~\ref{fig:largevpr}E). The non-monotonic behavior in some precision-recall curves, where precision temporarily improves with increasing recall, occured when the rate of true positive matches increased faster than false positives at certain threshold ranges. A spiking neural network to achieve this degree of accuracy with an architecture of just 753 neurons across three layers shows promise for large scale place recognition applications. We required the same amount of storage space capacity to store our model and the down sampled images for SAD, which was $\approx$179~KB.

\paragraph*{Localization performance during a navigation task}

\begin{figure}[H]
    \centering
    \includegraphics[width=0.70\textwidth]{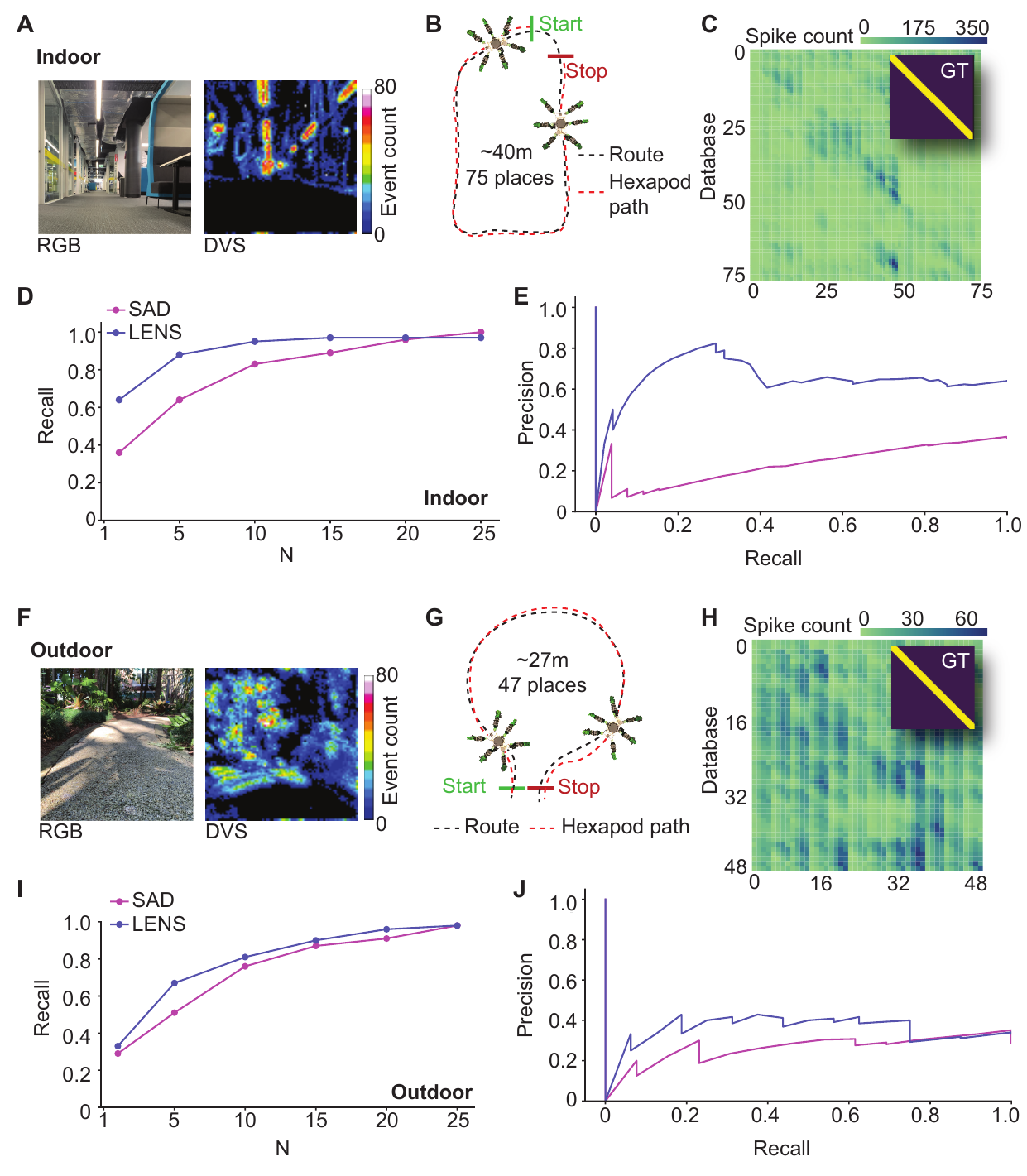}
    \caption{\textbf{Precision of a real-time neuromorphic robotic localization platform.} We deployed our system on a Hexapod robotic platform for on-device localization. (\textbf{A}) First, we tested an indoor environment by mapping and localizing along a predetermined route (\textbf{B}). As the Hexapod was teleoperated, there are slight differences in viewpoint between the mapping and localization phases. (\textbf{C}) To analyze the precision and recall of our indoor environment, we generated sequentially matched similarity matrices of the output spikes from all output neurons in our model while localizing. (\textbf{D}) Our compact system generally performed better than the sum of absolute differences (SAD) method for Recall@N, measuring accuracy in the top N matches, and Precision-Recall (\textbf{E}). We then deployed our Hexapod in an outdoor environment (\textbf{F}, \textbf{G}). The similarity matrix for the outdoor environment had more aliased incorrect matches than the indoor one (\textbf{H}), most likely due to the presence of multiple turns in the route. (\textbf{I, J}) In the outdoor task, LENS performed comparably to the SAD method in both the Recall@N and Precision-Recall evaluation.}
    \label{fig:matching}
    \end{figure}

To evaluate our system's real-time performance at the edge, we performed VPR on our hexapod in both indoor and outdoor environments (Fig.~\ref{fig:matching}). The model architecture was designed with $100\times200\times75$ for the input, feature, and output neurons, respectively. As illustrated in Figs.~\ref{fig:matching}A \& F, the generated event frames corresponded to the environment being traversed, particularly by detecting edges in the scene. The robot dealt with viewpoint and timing differences between the mapping and localization phases, as the route paths differed due to imperfect teleoperation (Figs.~\ref{fig:matching}B \& G). In comparison to the Brisbane Event Dataset, the hexapod represented a more unstable and challenging platform to perform place recognition.

For the indoor task, LENS achieved a Recall@1 of 0.64, outperforming SAD's performance at 0.36 (Fig.~\ref{fig:matching}C \& D). The precision-recall curve (PR-curve) indicated that LENS achieved higher precision at lower recall values compared to SAD (Fig.~\ref{fig:matching}E), suggesting that LENS was highly accurate in its identification of places. In the outdoor task, LENS performed comparably to SAD with a Recall@1 of 0.33 and 0.29, respectively (Fig.~\ref{fig:matching}H \& I). LENS also achieved a higher precision at lower recall, shown in the PR-curve (Fig.~\ref{fig:matching}J). The reduced performance in outdoor environments was likely due to the increased amount of turns in the trajectory, which caused widespread neuronal activation across the field of view. As indicated by the similarity matrix (Fig.~\ref{fig:matching}H), LENS showed a higher amount of activity across multiple output neurons. Fig.~\ref{fig:eventfreq} suggests that increased turning in the navigation route introduced noise, distributing spikes across more neurons. 

Overall, our compact spiking neural network model (130-150~KB in size) demonstrated comparable performance for VPR tasks in multiple environments, with key advantages of real-time deployability whilst being highly power and energy efficient. By contrast, the SAD method required 388~KB to store the images, $\approx$2.6 times the storage space needed for our model. Our system offered substantial benefits in terms of model size, making it feasible for deployment on computationally resource-constrained platforms. 

Interestingly, both SAD and LENS performed better on the Brisbane Event VPR dataset than during the on-robot VPR task (Fig.~\ref{fig:matching}). The Brisbane Event Dataset utilized a $346\times240$ sensor~\cite{Fischer2020}, compared to the $128\times128$ sensor~\cite{Lichtsteiner2008} on our robot. Additionally, the more stable platform (i.e., a car) used for dataset collection could reduce errors from viewpoint variance or swaying movement observed with the hexapod (see Supplementary Movie 1). These results indicate that our proposed model can be effectively deployed for both short and long-range place recognition tasks for robotic localization.  

\section*{Discussion}

This work demonstrated a fully neuromorphic visual place recognition system capable of real-time localization, which was successfully deployed on a hexapod robotic platform without any external sensing or computational resources. The LENS spiking neural network model, characterized by its simplicity and compactness, performed accurately across multiple environments and scales, demonstrating its versatility for a range of navigation platforms and tasks. Our system integrated with the Speck\texttrademark{} chip operated with 1-2 orders of magnitude lower energy than conventional von Neumann hardware, underscoring its viability for deployment on resource-constrained platforms such as unmanned aerial vehicles~\cite{Bian2023}, underwater~\cite{Kelasidi2015}, and locomotive robotics~\cite{Kashiri2018}.

Notably, our models were less than 180~KB in size, with the largest containing just 44~K parameters. Accurate localization with our compact model enables portable deployment on neuromorphic processors with higher memory capability for larger models to improve overall performance (Fig.~\ref{fig:modelsize}). Previous research has shown that accurate place recognition with event cameras can be achieved with as few as 25 pixels~\cite{Fischer2022}. Our system capitalized on this by balancing the number of pixels with the size of the feature layer to achieve the optimal performance (Fig.~\ref{fig:modelsize}). A limiting factor in this work was the memory capacity of the neurocores for Speck\texttrademark{}, which cap at 64~KB per core. Overcoming this limitation with neuromorphic hardware offering higher memory capabilities could further enhance the performance of large-scale place recognition tasks (Fig.~\ref{fig:modelsize}). Nonetheless, we have shown comparable performance for place recognition in highly compact spiking network models across multiple environments.

Our neuromorphic place recognition system employs real event streams directly for localization in a unique way, establishing a new methodology for event-driven place recognition. This work highlights the potential for substantial efficiency gains when using event-based DVS over conventional cameras, especially in scenarios where energy-efficient processing is critical. Unlike frame based cameras, which operate continuously, the event-driven DVS captures only essential changes, reducing data redundancy and energy consumption. Our experiments encompass both a conventional benchmark dataset widely used in the robotics community~\cite{Fischer2020} and a more challenging legged robot navigation task, ensuring that the localization accuracy of our method is evaluated across a diverse range of environments and baseline performance levels. The accuracy achieved by LENS demonstrates its effectiveness within broader systems such as Simultaneous Localization and Mapping (SLAM), where geometric constraints and filtering methods can mitigate occasional inaccuracies~\cite{Islam2021, Mur-Artal2015, Yang2022}. By utilizing a more stable platform, such as a four wheeled robot, our network is able to achieve high precision comparable to a benchmark technique. This demonstrates that our compact spiking network models perform comparably to conventional non-event-driven methods, highlighting the potential for further improvements with future developments in neuromorphic hardware and algorithmic enhancements. 

As observed in the similarity matrices for the localization tasks (Fig.~\ref{fig:matching}), introducing many turns into shorter trajectories proved challenging for LENS in the outdoor hexapod experiment. Dynamically adjusting these biases and parameters of DVS event streams to better handle these sudden motions could enhance accuracy across challenging environments~\cite{Nair2024}. Continued development of this system will focus on detecting and handling turning motions to improve localization performance. 

Another limitation of our system is the use of static, temporal representations of places generated from event streams using time-window binning for model training (Fig.~\ref{fig:lensschema}). Training our model directly on DVS event streams could help overcome the localization challenges observed in dynamic environments. Developing this would also facilitate online learning capabilities, where event streams are used directly to train network models in real-time. This would enable the system to adapt to new environments on-the-go, potentially incorporating introspective capabilities that allow the robot to recognize when it is uncertain about its location and learn accordingly~\cite{Wu2022,Zhu2023,Amirhossein2022,Sangwoo2023}. Additionally, using more information from the DVS sensor such as including convolutional layers and self-supervised learning could be useful to extract more features for enhanced performance~\cite{Chen2020}.

Integration of our neuromorphic system with other methods for navigation could be used to enhance and improve robotic localization performance. Dupeyroux \textit{et al.} developed a bio-inspired path integration system for a hexapod robotic platform that would serve as a complementary method for VPR~\cite{Dupeyroux2019}. Additionally, other insect inspired systems employ methods such as goal approaching and collision avoidance which allows for autonomous navigation in novel environments, typically for homing tasks~\cite{Dupeyroux2019, SUN2023, vanDijk2024}. Integration of these mixed-model paradigms present themselves as a promising method to be used in more comprehensive navigation systems such as SLAM, of which there are many neuroscience inspired systems available~\cite{Milford2004-2,Dumont2023, Yu2019, Zeng202021}.

In conclusion, we present a fully end-to-end neuromorphic VPR system that performs accurate, low-power localization on robotic platforms. Our work has the potential to enable robots in the field to navigate further and conduct longer missions by reducing the computational and energy demands of current vision systems~\cite{liu2019}. An important aspect of this work was to ensure that neuromorphic localization could be achieved on multiple robotic platforms, including for resource-constrained scenarios. Our approach moves VPR toward efficient, versatile neuromorphic localization, unlocking new possibilities for autonomous robotic navigation at the edge.

\section*{Materials and Methods}

\paragraph*{Model training}

Our network models were trained on CPU hardware or a Jetson Nano using static temporal representations of event streams with pixel intensity values relating to the number of events collected over a one second time-window, subsequently normalized in the range $[0, 1]$ (see Fig.~\ref{fig:lensschema}). The network was trained using spike timing from an abstracted spiking method, where the pixel intensity value represents temporal activity~\cite{Hines2024,Stratton2022, Guo2021}. The networks three layer architecture (input, feature, and output) used unsupervised learning between the input and feature layers, and a supervised delta learning rule for the feature to output layer.

Unsupervised learning applied spike-timing dependent plasticity (STDP) to update weights $W$, encouraging or pruning connections between neurons. Initial weights and connections were randomly generated from predetermined probabilities, with connections either being positive (\emph{excitatory connection}) or negative (\emph{inhibitory connection}). From the input to feature layer there was a $35\%$ and $75\%$ excitatory and inhibitory connection probability, respectively. The feature to output layer was fully connected with an equal distribution of excitatory and inhibitory connections. As spikes were propagated through the network, the layer weights were updated in the following way:

    \begin{align}
    \begin{gathered}
        \Delta W^{nm}_{ji}(t) = \frac{\eta_\text{STDP}(t)}{f^n_j} \cdot \Big[\Theta\big(x^m_i(t-1)\big)\cdot \Theta\big(x^n_j(t)\big)\cdot \big(0.5-x^n_j(t)\big)\Big],
    \label{eq:weightupdate}
    \end{gathered}
    \end{align}
    where $W^{nm}_{ij}$ is the connection weight between neuron $j$ in layer $n$ to neuron $i$ in layer $m$, $\eta_\text{STDP}$ is the STDP learning rate, $f_j^n$ is the target firing rate of neuron $j$ in layer $n$, $\Theta(\cdot)$ is the Heaviside step function, $x^m_i$ and $x^n_j$ are the neuron states $x$ in the network layer $i$ and $j$ with connections $m$ and $n$, and $t$ is the timestep. During weight updates throughout training, we did not specify any rule that says weights must remain excitatory or inhibitory, however we did not observe weights switching sign. That is, initially seeded connection types remained robust throughout training.

For supervised learning between the feature and output layer, we employed a delta learning rule to force each output neuron to learn the feature representation of a single reference place~\cite{Hines2024,Stratton2022}:

    \begin{align}
    \begin{gathered}
        \Delta W^{nm}_{ji}(t) = \eta_\text{STDP}(t)/f^n_j . [x^m_i(t-1)(x_\text{force j}^n(t) - x^n_j(t))],
    \label{eq:spikeforce}
    \end{gathered}
    \end{align}
    where $x_\text{force}$ was the forced spike in the selected output neuron during training.
    
    The hyperparameters used for our models can be found in Table~\ref{tab:params} which were obtained by performing a hyperparameter sweep.

\paragraph*{Deployment time}

The Locational Encoding with Neuromorphic Systems (LENS) framework leveraged spiking neural networks (SNN) on neuromorphic processors for computationally and energy efficient place recognition. We deployed our pre-trained SNN model on a SynSense Speck2fDevKit, which incorporates the Speck\texttrademark{} neuromorphic processor with a $128\times128$ dynamic vision sensor (DVS)~\cite{Yao2024} for event driven, real-time localization. This was in contrast to the training methodology where event frames were generated by counting the number of events over a specified time-window. Our localization system used truly asynchronous event streams to continually bias and activate neurons as the robot navigated through the environment.

To optimize input size and remove low-activity pixels, we first selected an $80\times80$ region of interest (ROI) in the top-center of the sensor. Events from this ROI were processed using a 2D convolution layer with a kernel size and stride of 8, which acted as the first processing layer of our SNN prior to the pre-trained input layer and was deployed on chip. We set the center weight of the kernel to 1, with all other weights at 0, effectively selecting the center pixel from each convolution to reduce the input to 100 neurons. This selection approach followed a previously established method for event pixel selection~\cite{Fischer2022}. 

The input neurons were sparsely connected to a linear feature layer of 200 neurons, with excitatory and inhibitory connection probabilities set at $35\%$ and $75\%$ respectively. The feature layer was fully connected to the output layer, with the number of output neurons corresponding to the number of learned places (see Materials and Methods: Model training). Events were asynchronously collected and passed into our model, where Constant-leak Integrate and Fire (IAF) neurons propagated spikes across the network layers:

    \begin{align}
    \begin{gathered}
        \tau \dot{v} = -v_\text{leak} + R \cdot (I_\text{syn} + I_\text{bias}),
    \label{eq:IAFmodel}
    \end{gathered}
    \end{align}
    where $\tau$ is the membrane time constant, $v$ is the membrane potential, $R$ is the constant resistance value used to match units between membrane potential and currents, $I_\text{syn}$ is the weighted sum of all input synaptic contributions, and $I_\text{bias}$ is a constant bias.

\paragraph*{Online place matching}

Place matching was performed using real-time sequence matching~\cite{Milford2012, Garg2021-2} to enhance the precision of our VPR model on the Speck\texttrademark{}. Spike counts from the output layer were collected in one second time bins, then averaged over four time bins (four seconds), representing $\approx$0.5$\,\text{m}$ of robot movement (average speed of 13.5~$\text{cm}\cdot\text{s}^{-1}$). All place and sequence matching was performed off-chip on a Jetson Nano during on-robot deployment or a CPU for simulated experiments. Once four sets of output spike counts were gathered, representing about 2~m of travel, they were compiled into a distance matrix $M_{dist}$ (Fig.~\ref{fig:schemafig}D), which was convolved ($*$) with an identity matrix $\mathbb{I}_L$ to generate a sequence-based distance matrix $M_{seq}$, as previously described~\cite{Garg2021-2}:  

    \begin{align}
    \begin{gathered}
        M_{seq}(i, j) =  \mathbb{I}_L * M_{dist}(i, j) / L,
    \label{eq:seqmatch}
    \end{gathered}
    \end{align}
    where $i$ and $j$ are the row and column of the matrix, respectively, and $L$ is the sequence length, which we set to 4 matching the number of output spike counts. For the Brisbane Event Dataset, we set the sequence length to 30. The choice of sequence length matters for both computational efficiency and evaluation performance, with shorter lengths favoring the former and longer lengths benefiting the latter (Fig.~\ref{fig:seqlengthablation}).

After generating the sequence-based distance matrix $M_{\text{seq}}$, we performed place matching on the processed spike outputs. The matched place $\hat{p}$ was identified as the place $i$ corresponding to the neuron with the highest spike count $x_i$ for each query in the $M_{\text{seq}}$:

    \begin{align} 
    \begin{gathered}
        \hat{p}_k = \arg\max_{i} M_{\text{seq}}(k, i) \quad \text{for each} \quad k \in {1, 2, 3, 4}.
    \label{eq:placematch} 
    \end{gathered}
    \end{align}
    where $\hat{p}_k$ is the matched place for the $k$-th sequence, $i$ ranged over all database reference places, $M_{\text{seq}}(k,i)$ represents the sequence-based distance between the $k$-th spike count set and the $i$-th reference place. 
    
\paragraph*{Off-chip simulation of event streams}

For power and energy consumption comparisons using LENS on neuromorphic and von Neumann hardware, we simulated event-driven localization using event frames from the Speck\texttrademark{} energy experiments (see Materials and Methods: Power and energy measurements). Integrate And Fire neuron spike rates were generated using a time-based rate code: 
    
    \begin{align}
    \begin{gathered}
        I_{t,h,w} = \begin{cases} 
        1 & \text{if } R_{t,h,w} < X_{h,w} \\
        0 & \text{otherwise},
        \end{cases}
        \label{eq:ratecodes}
    \end{gathered} 
    \end{align}
    where $I_{t,h,w}$ is a boolean tensor with spatial dimensions $h$ and $w$ over $t$ time-windows, where $1=\text{spike}$ and $0=\text{no spike}$. $R_{t,h,w} \in [0, 1]$ is a uniformly random tensor, and $X_{h,w}$ is the original input. Localization was simulated over 1000 time windows with $dt=0.001s$ to represent 1$s$ of output spike collection, similar to the online place matching (see Materials and Methods: Online place matching).

\paragraph*{Robotic deployment}

We used the commercially available JetHexa hexapod (Hiwonder, Shenzen China) customized with a 3D-printed mount to house the Speck2fDevKit (see Fig.~\ref{fig:schemafig}). We also simulated a four wheeled robotic deployment by evaluating our system on the Brisbane Event Dataset~\cite{Fischer2020}, to highlight that our energy-efficiency performance is robotic platform agnostic. Training data was collected by teleoperating the robot for a reference traversal. Real-time localization was performed on the Speck\texttrademark{} neuromorphic processor with the pre-trained model closely following the reference traversal. Indoor and outdoor datasets represented approximately 40~m and 27~m of traversal, encoding 75 and 48 unique places, respectively. To ensure accurate ground truth alignment between the mapping and localization phases, experiments were consistently started and ended at the same location.

\paragraph*{Power and energy measurements}

Intel\textsuperscript{\textregistered} CPU power was measured using SoC Watch, and NVIDIA Jetson Nano power was recorded with the jetson-stats Python package, with baseline idle power subtracted in both cases to only measure energy consumed for the VPR algorithms. Speck\texttrademark{} power tracks (IO, RAM, memory, VDDA, and VDDD) were recorded at 20 Hz using SynSense's samna package during testing (see Supplementary Movie 1 and Fig.~\ref{fig:energy}).

\paragraph*{Statistical analysis}

We evaluated our system using Recall@N and precision-recall curves, two standard metrics for VPR studies~\cite{Schubert2023}. All statistical analysis was performed on CPU hardware. Our implementation of precision and recall matched methods used in other place recognition systems~\cite{Yu2023} (Fig.~\ref{fig:prcurvecomp}). Recall@N measured the accuracy of the system such that for $N=1$ only the highest match was considered across all reference places, whereas for $N=5$ the top 5 matches were considered when calculating the accuracy, up to a maximum of 25 potential places. Precision-recall curves were utilized to assess performance by illustrating the relationship between precision and recall across different threshold levels for positive matches, calculated as follows: 

    \begin{align}
    \begin{gathered}
        Precision = \frac{TP}{TP + FP} \hskip1em
        Recall = \frac{TP}{GTP}
    \label{eq:precrecall}
    \end{gathered}
    \end{align}
    where $TP$ is the number of true positives, $FP$ is the number of false positives, and $GTP$ is the number of ground truth positives~\cite{Schubert2023}. Initially precision and recall is undefined due to there being no predictions, however we initialize precision at 1 and recall at 0 for visualization purposes.
    
We used the Sum of Absolute Differences (SAD) method for comparison, which computed pixel-wise similarity across reference and query images and has been used previously for place recognition using event streams~\cite{Fischer2022,Milford2012}. Pseudo ground-truth data was collected following~\cite{Fischer2022} using GPS correspondence between reference and query images. Pseudo-GPS and manually confirmed ground-truth showed high correspondence, with the majority of places being ± 1 place label away (Fig.~\ref{fig:pseudomanualgt}). Further analysis of the accuracy of the pseudo ground-truth and applied tolerances were confirmed by near identical results (Fig.~\ref{fig:pseudomanualpr}). We allowed a ground-truth tolerance of $\approx$1~m, which is equivalent to $\pm$ 2 reference places, and a tolerance of $\approx$42~m for the Brisbane Event VPR dataset, which is equivalent to $\pm$ 3 reference places. A visualization of the ground truth, TP, and FP matches for each similarity matrix can be found in Fig.~\ref{fig:gttpfp}. 

\paragraph*{Data and code availability}

All data needed to support the conclusions of this manuscript are included in the main text, Supplementary Materials, and are also available alongside the code at \url{https://doi.org/10.5281/zenodo.15392412}.

\newpage

\clearpage %
\bibliography{scibib}
\bibliographystyle{sciencemag}

\section*{Acknowledgments}
The authors acknowledge continued support from the Queensland University of Technology (QUT) through the Centre for Robotics. We wish to acknowledge the support of the Research Engineering Facility (REF) team at QUT for the provision of engineering support, expertise and research infrastructure in enablement of this project. We also would like to thank Dr Scarlett Raine and Dr Somayeh Hussaini for their valuable feedback during initial manuscript preparation. Finally, we would like to thank the organisers and participants of the 2022 Lifelong Learning at Scale topic area at the Telluride Neuromorphic Workshop for the insightful discussions and inspiring environment, which helped shape some of the ideas explored in this paper. \paragraph*{Funding:} This work received funding from an ARC Laureate Fellowship FL210100156 to MM, AUSMURIB000001 associated with ONR MURI grant \mbox{N00014191-2571} to MM and TF, and an ARC Discovery Early Career Researcher Award DE240100149 to TF.  \paragraph*{Author contributions:} 
A.D.H \& T.F. conceptualized and designed the experiments; A.D.H performed and analyzed all experiments; T.F. \& M.M. provided general input into the research; T.F. \& M.M. provided funding for the project; A.D.H wrote the paper; A.D.H., M.M. \& T.F. edited the paper. \paragraph*{Competing interests:} The authors declare no competing interests. \paragraph*{Data and materials availability:} All data and code used in this study is available at \url{https://doi.org/10.5281/zenodo.15392412}.
\clearpage

\subsection*{Supplementary materials}
Supplementary Figures S1-S8\\
Supplementary Tables S1-S2\\
Supplementary Movie S1

\newpage

\renewcommand{\thefigure}{S\arabic{figure}}
\renewcommand{\thetable}{S\arabic{table}}
\renewcommand{\theequation}{S\arabic{equation}}
\renewcommand{\thepage}{S\arabic{page}}
\setcounter{figure}{0}
\setcounter{table}{0}
\setcounter{equation}{0}
\setcounter{page}{1} %

\begin{center}
\section*{Supplementary Materials for\\ \scititle}
Adam D Hines$^{1\ast}$, Michael Milford$^{1}$, Tobias Fischer$^{1}$\\
\small{$^{1}$QUT Centre for Robotics, Queensland University of Technology, 2 George St, Brisbane, QLD 4001, Australia}\\
\small{$^\ast$Corresponding author. Email:  adam.hines@qut.edu.au}
\end{center}

\subsubsection*{This PDF file includes:}
Supplementary Figures S1-S8\\
Supplementary Tables S1-S2\\

\subsubsection*{Other Supplementary Materials for this manuscript:}
Supplementary Movie S1

\newpage

\begin{figure}[ht]
    \centering %
    \includegraphics[width=\textwidth]{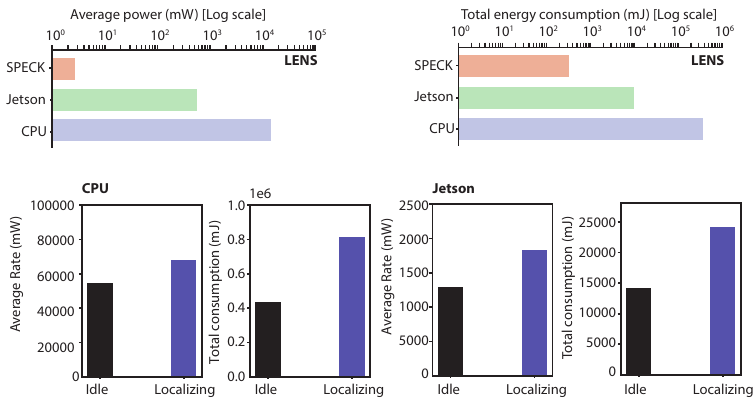} %
    \caption{\textbf{Power and energy measurements for running LENS on von Neumann hardware.} The baseline power consumption and energy was subtracted from the localization measurements for our comparisons to other hardware.} 
    \label{fig:lensvonneu}
\end{figure}

\begin{figure}
    \centering %
    \includegraphics[width=\textwidth]{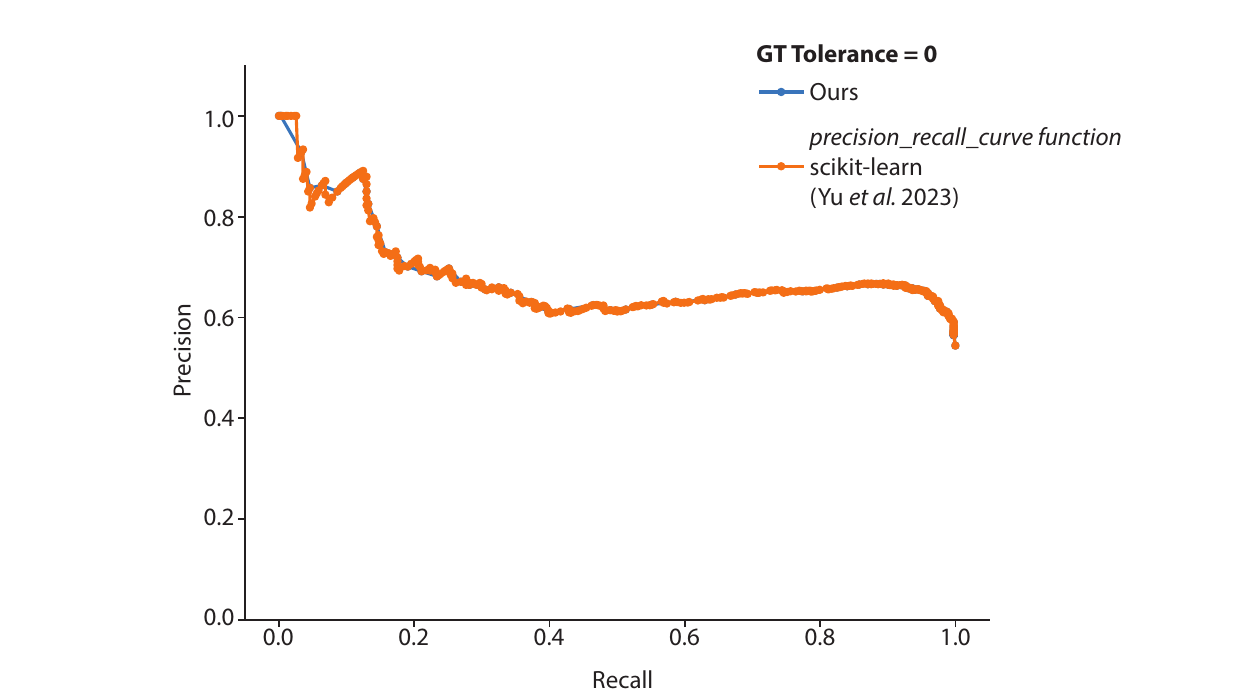} %
    \caption{\textbf{Comparison of Precision-Recall calculation without ground truth tolerance.} PR curves for the Brisbane Event-VPR dataset generated by our pipeline (blue) align closely with those generated using the scikit-learn $\text{precision\_recall\_curve}$ function (orange) employed by Yu et al.~\cite{Yu2023}. The minor deviation is due to discretization artefacts, as our method uses 100 points while the approach taken by Yu et al. uses 398 points to compute the curves.} 
    \label{fig:prcurvecomp}
\end{figure}

\begin{figure}
    \centering %
    \includegraphics[width=0.79\textwidth]{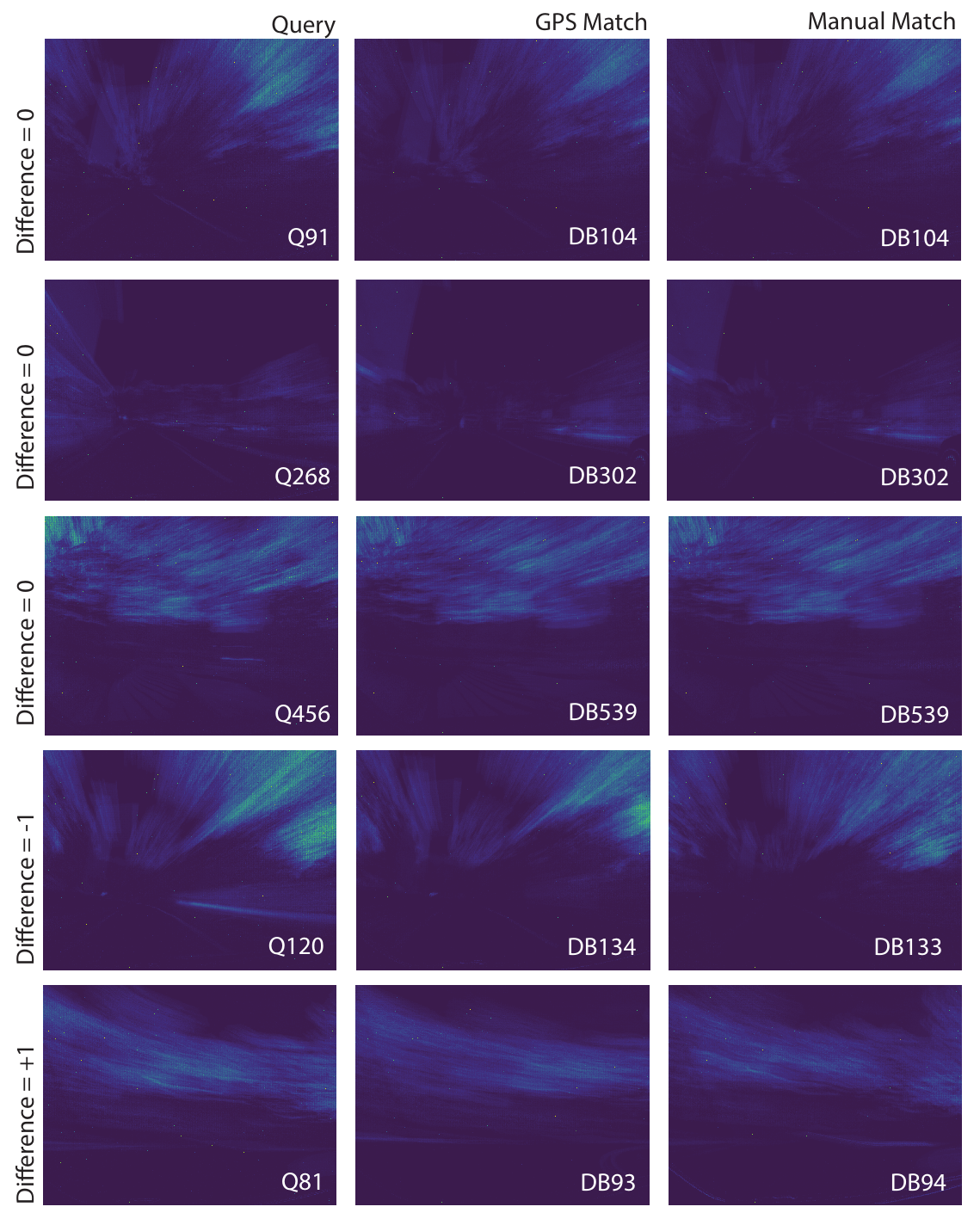} %
    \caption{\textbf{Comparison of GPS and human-labeled ground truth for queries and their matching references.} (Left) Shows the query images with an inset of their place index for the Brisbane Event-VPR dataset. (Middle) The matching reference place using the GPS pseudo ground truth and the (Right) actual match confirmed by manual ground truth. In the cases where Difference = 0, the GPS pseudo ground truth matches exactly to the manually labeled correct image. In cases where Difference=+/- 1, the GPS match was offset by one image of the true match, which is equivalent to approximately 14~m. The purpose of the ground truth tolerances is to allow such neighboring images to count as correct matches, accounting for such slight misalignments.} 
    \label{fig:pseudomanualgt}
\end{figure}

\begin{figure}
    \centering %
    \includegraphics[width=0.75\textwidth]{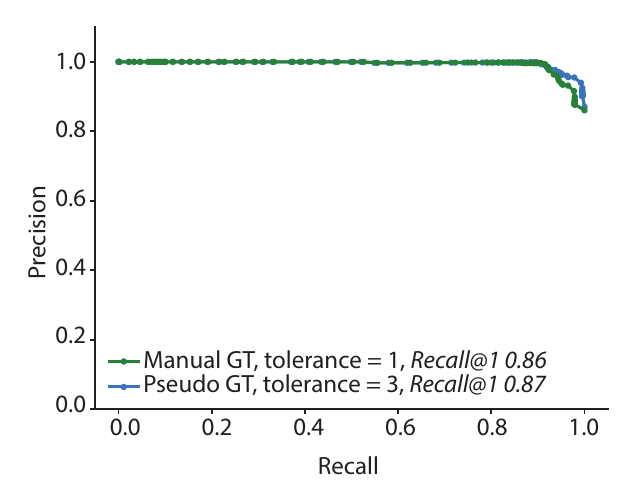} %
    \caption{\textbf{Precision-recall evaluation comparing GPS pseudo ground truth with manually aligned ground truth.} For the Brisbane Event-VPR dataset, we compare the GPS pseudo ground truth (GT) with the manually annotated ground truth using tolerances of 1 and 3 places representing approximately 14~m and 42~m, respectively. Both performed very similarly with a Recall@1 of 0.86 and 0.87 for the manual and GPS pseudo ground truth, respectively. The slight improvement in performance observed with the pseudo ground truth can be attributed to the higher tolerance threshold of 42~m capturing 7 additional matches, which extends beyond the 14~m limit used for manual ground truth correction.} 
    \label{fig:pseudomanualpr}
\end{figure}

\begin{figure}
    \centering %
    \includegraphics[width=0.85\textwidth]{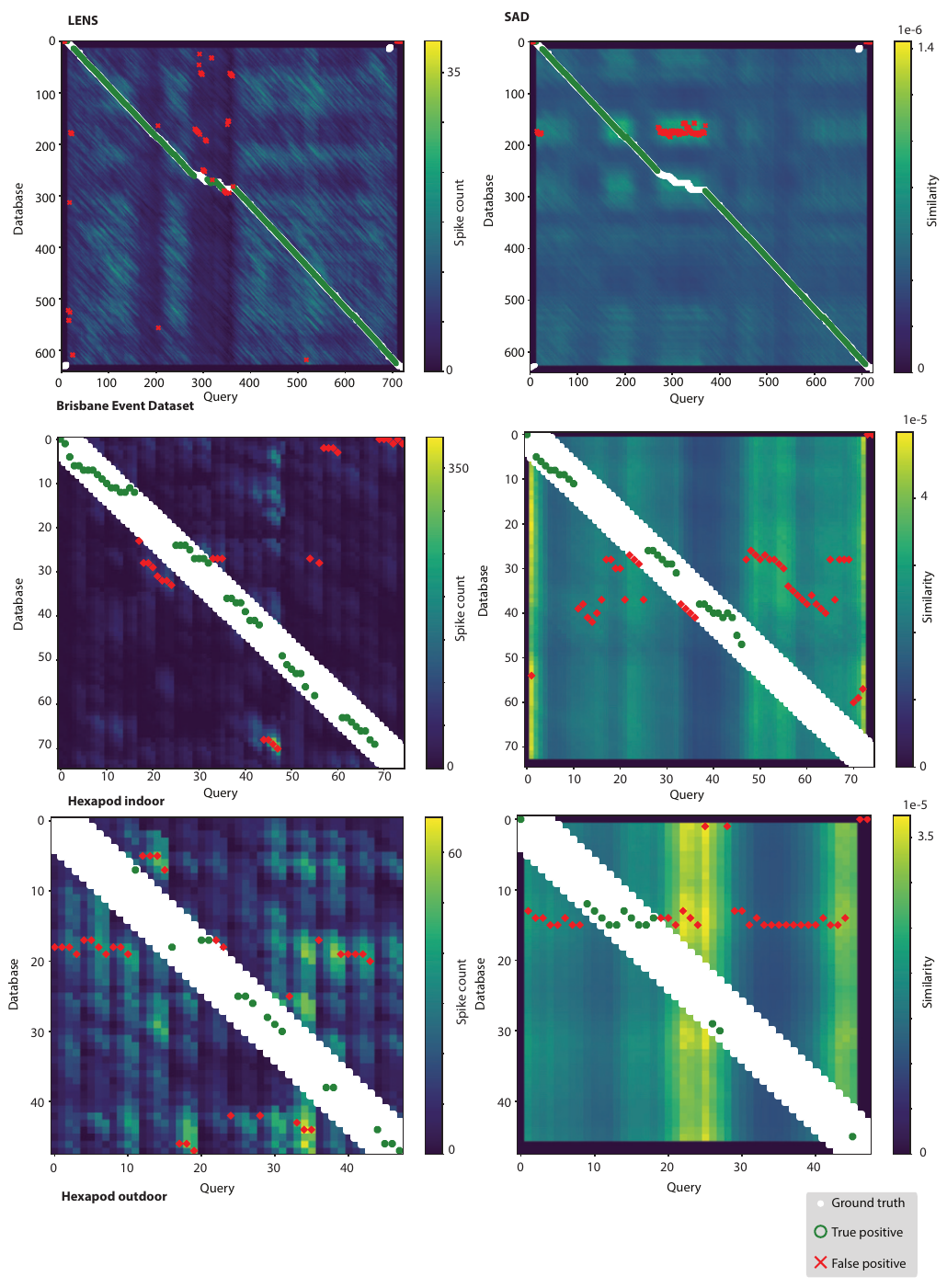} %
    \caption{\textbf{Distribution of true and false positives for similarity matrices.} Similarity matrices for (left) LENS and (right) SAD for the 3 datasets evaluated from top to bottom - Brisbane Event Dataset, hexapod indoor, and hexapod outdoor. Matrices show the ground truth used (white dots) with true positives (green circles) and false negatives (red crosses).}
    \label{fig:gttpfp}
\end{figure}

\begin{figure}
    \vspace{-2cm} %
    \centering %
    \includegraphics[width=0.68\textwidth]{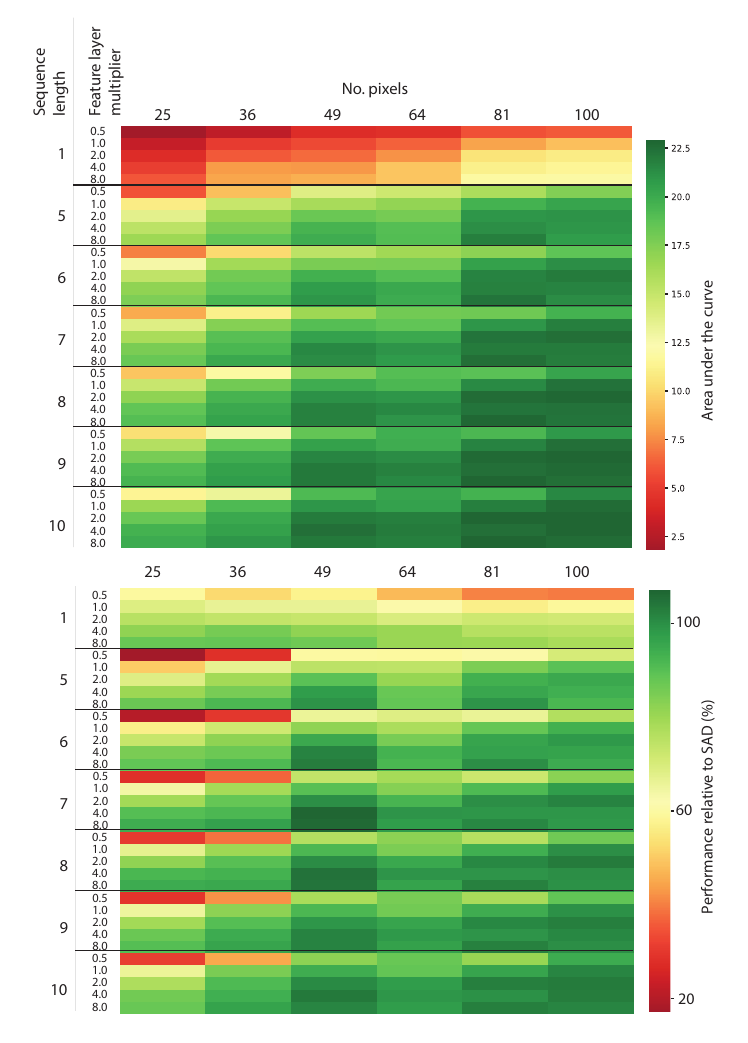} %
    \caption{\textbf{Effects of model size and sequence length on network performance.} (\textbf{Top}) Heatmap of the area under the curve of the Recall@N results from training and testing the Brisbane Event VPR data~\cite{Fischer2020} for different number of input pixels, feature layer size multiplier (relative to number of pixels), and sequence lengths. Our model with $49\times49\times641$ neurons with a sequence length of 10 was the highest performer whilst still being able to be deployed to Speck\texttrademark{}. The best performing model had the architecture of $100\times200\times641$ neurons and a sequence length of 10, indicating that model accuracy heavily depends on the number of input events the ratio of the feature layer to the input. (\textbf{Bottom}) Performance of the various models relative to sum of absolute differences for the area under the curve. Smaller number of input pixels with larger feature layer sizes can perform as well as or better than a larger number of input pixels with the same feature layer size. The important consideration is that increasing the feature layer size greatly increases the model size, which may limit deployment on memory constrained hardware.}
    \label{fig:modelsize}
\end{figure}

\begin{figure}
    \centering %
    \includegraphics[width=\textwidth]{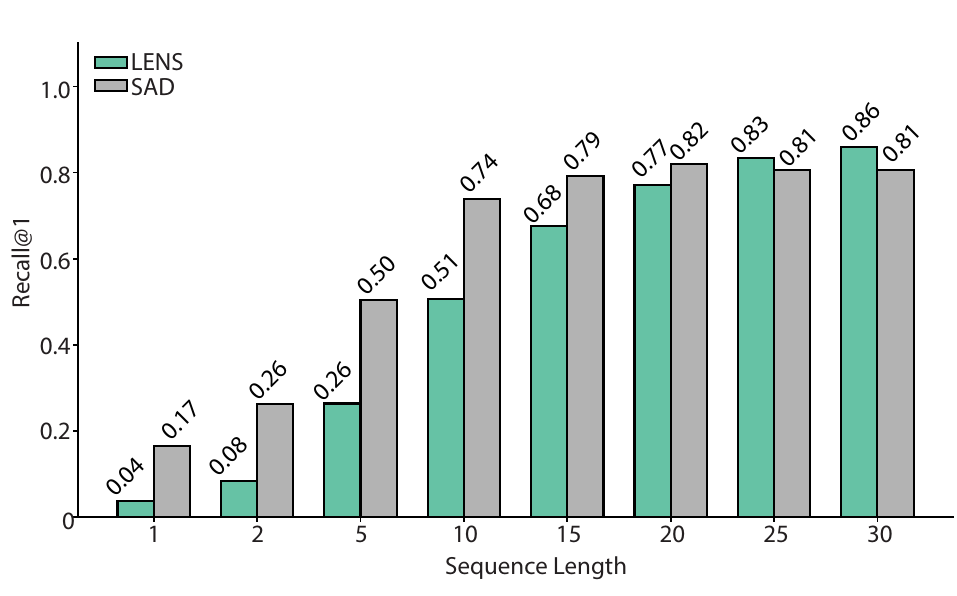} %
    \caption{\textbf{Effect of sequence length on Recall@1 performance.} An ablation study was conducted on the Brisbane Event-VPR dataset observing the effects of sequence length for Recall@1 performance between our system (LENS) and the Sum-of-Absolute-Differences (SAD) baseline method. We used the manually generated ground truth with a tolerance of +/- 1 place (approximately 14 m) for this investigation.}
    \label{fig:seqlengthablation}
\end{figure}

\begin{figure}
    \centering %
    \includegraphics[width=0.8\textwidth]{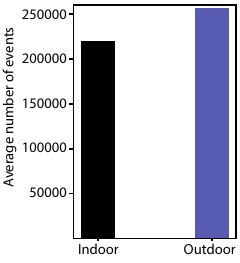} %
    \caption{\textbf{Average number of events between indoor and outdoor traversals.}}
    \label{fig:eventfreq}
\end{figure}

\clearpage

\begin{table}[h]
    \caption{Hyperparameters for network training}
    \centering
    \renewcommand{\arraystretch}{1.2}

    \begin{tabular}{c|c|c|c|c|c|c}
        \hline
        \textbf{Parameter} & \textbf{$\boldsymbol{\theta_{\text{max}}^{I\rightarrow F}}$} & \textbf{$\boldsymbol{\theta_{\text{max}}^{F\rightarrow O}}$} & \textbf{$\boldsymbol{\eta^{\text{init}}_{\text{STDP}}}$} & \textbf{$\boldsymbol{\eta^{\text{init}}_{\text{ITP}}}$} & \textbf{$\boldsymbol{f_{\text{min}}^{I\rightarrow F}, f_{\text{max}}^{I\rightarrow F}}$} & \textbf{$\boldsymbol{f_{\text{min}}^{F\rightarrow O}, f_{\text{max}}^{F\rightarrow O}}$} \\
        \hline
        \textbf{Values} & 0.75 & 0.5 & 0.01 & 0.02 & [0.4, 0.6] & [0.5, 0.5] \\
        \hline
        \textbf{Parameter} & \textbf{$\boldsymbol{P_{\text{exc}}^{I\rightarrow F}}$} & \textbf{$\boldsymbol{P_{\text{inh}}^{I\rightarrow F}}$} & \textbf{$\boldsymbol{P_{\text{exc}}^{F\rightarrow O}}$} & \textbf{$\boldsymbol{P_{\text{inh}}^{F\rightarrow O}}$} & \textbf{$\boldsymbol{\epsilon^{I\rightarrow F}}$} & \textbf{$\boldsymbol{\epsilon^{F\rightarrow O}}$} \\
        \hline
        \textbf{Values} & 0.35 & 0.75 & 1.0 & 1.0 & 64 & 128 \\
        \hline
    \end{tabular}
    \label{tab:params}
\end{table}

\begin{table}[h]
    \caption{Energy consumption for SAD with fewer pixels}
    \centering
    \renewcommand{\arraystretch}{1.2}

    \begin{tabular}{c|c|c|c}
        \hline
        \textbf{Pixel count} & \textbf{100} & \textbf{64} & \textbf{1} \\
        \hline
        \textbf{Energy CPU (mJ)} & 61,427.0 & 46943.18 & 25698.37 \\
        \hline
        \textbf{Energy Jetson (mJ)} & 2,968.0 & 531.81 & 474.88 \\
        \hline
    \end{tabular}
    \label{tab:sadenergy}
\end{table}

\clearpage %
\paragraph{Caption for Supplementary Movie S1.}
\textbf{LENS display during localization.} The left side shows the main display of LENS during online localization deployed on a robotic platform. The Visualizer shows the field of view of the on-board dynamic vision sensor with the power being actively consumed dynamically plotted underneath. The right side shows the output of the LENS matching system which returns the place matching result after collecting output spikes from the model over the specific time window.

\end{document}